%% file: aaai.tex
\documentclass[letterpaper]{article} % DO NOT CHANGE THIS
\usepackage{aaai23}  % DO NOT CHANGE THIS
\usepackage{times}  % DO NOT CHANGE THIS
\usepackage{helvet}  % DO NOT CHANGE THIS
\usepackage{courier}  % DO NOT CHANGE THIS
\usepackage[hyphens]{url}  % DO NOT CHANGE THIS
\usepackage{graphicx} % DO NOT CHANGE THIS
\usepackage{natbib}  % DO NOT CHANGE THIS AND DO NOT ADD ANY OPTIONS TO IT
\usepackage{caption} % DO NOT CHANGE THIS AND DO NOT ADD ANY OPTIONS TO IT
\usepackage{algorithm}
\usepackage{algorithmic}
\usepackage{commath}
\usepackage{amsmath}
\usepackage{amssymb}
\usepackage{mathtools}
\usepackage{amsthm}
\usepackage{booktabs}

\usepackage{newfloat}
\usepackage{listings}
\DeclareCaptionStyle{ruled}{labelfont=normalfont,labelsep=colon,strut=off} % DO NOT CHANGE THIS
\floatstyle{ruled}
\newfloat{listing}{tb}{lst}{}
\floatname{listing}{Listing}
\renewcommand{\SS}{\mathbb{S}}
\renewcommand{\AA}{\mathbb{A}}
\newcommand{\PP}{\mathcal{P}}
\newcommand{\RR}{\mathbb{R}}

\title{The Sufficiency of Off-Policyness and Soft Clipping: \\
PPO is still Insufficient according to an Off-Policy Measure}
\author {
    Xing Chen,\textsuperscript{\rm 1,4}
    Dongcui Diao, \textsuperscript{\rm 6}
    Hechang Chen, \textsuperscript{\rm 1,4,*}
   Hengshuai Yao,\textsuperscript{\rm 2,*}
    Haiyin Piao, \textsuperscript{\rm 3}
    Zhixiao Sun, \textsuperscript{\rm 3}
    Zhiwei Yang, \textsuperscript{\rm 1,4}  
   Randy Goebel, \textsuperscript{\rm 2,5}
   Bei Jiang, \textsuperscript{\rm 5,6}
   Yi Chang,\textsuperscript{\rm 1,4,*}
}
\affiliations {
    \textsuperscript{\rm 1} School of Artificial Intelligence, Jilin University, Changchun, China\\
    \textsuperscript{\rm 2} Department of Computing Science, University of Alberta, Edmonton, Canada\\
     \textsuperscript{\rm 3} School of Electronics and Information, Northwestern Polytechnical University, Xian, China\\
     \textsuperscript{\rm 4} Engineering Research Center of Knowledge-Driven Human-Machine Intelligence, Ministry of Education, China\\
     \textsuperscript{\rm 5} Alberta Machine Intelligence Institute, University of Alberta, Edmonton, Canada\\
     \textsuperscript{\rm 6} Department of Mathematical and Statistical Sciences, University of Alberta, Edmonton, Canada\\
    xingchen19@mails.jlu.edu.cn, \{chenhc,yichang\}@jlu.edu.cn, hengshu1@ualberta.ca
}

\begin{document}

\maketitle

\begin{abstract}
%Many policy gradient methods seek to optimize the Conservative Policy Iteration (CPI) objective, which involves importance sampling (IS) ratios. Optimizing this objective is problematic because extremely large IS ratios can cause algorithms to catastrophically fail. 
The popular Proximal Policy Optimization (PPO) algorithm approximates the solution in a clipped policy space. Does there exist better policies outside of this space? By using a novel surrogate objective that employs the sigmoid function (which provides an interesting way of exploration), we found that the answer is ``YES'', and the better policies are in fact located very far from the clipped space. We show that PPO is insufficient in ``off-policyness'', according to an off-policy metric called DEON. Our algorithm explores in a much larger policy space than PPO, and it maximizes the Conservative Policy Iteration (CPI) objective better than PPO during training. To the best of our knowledge, all current PPO methods have the clipping operation and optimize in the clipped policy space. Our method is the first of this kind, which advances the understanding of CPI optimization and policy gradient methods. Code is available at https://github.com/raincchio/P3O.
\end{abstract}

\section{Introduction}
Real-world problems like medication dosing and autonomous driving pose a great challenge for Artificial Intelligence (AI) with an expectation to improve human life and safety. Applications like these require significant interaction with the environment to make learning algorithms effective. Humans can learn from others, by observing their experience to quickly pick up new skills, even without exposure on one's own. Subsequently, when there is a chance to practice, the skills obtained from previous experience or others can be quickly adapted and improved.
But it is clear that we are still far from obtaining this remarkable learning ability in AI.

In our context, we consider problems where the environment state cannot be reset, which is also true in real life. In such situations, we can only sample a limited number of possible trajectories, which easily results in failure of learning.

Off-policy learning is one promising paradigm to address this challenge, and it provides an effective discipline of learning by sampling the potential trajectories starting from any state. This means we can evaluate a target policy using a behavior policy that generates experience \citep{offpolicy_doina,gtd}. Moreover, the paradigm seems characteristic and general enough for obtaining skills from other sources. With off-policy learning, one agent can reuse experience from itself or even the other agents, where the samples are collected with methods that are different from real-time on-policy interaction. Off-policy learning holds significant promise, but it is tricky in practice. The mismatch between the distribution of the behavior policy and that of the target policy poses a big stability challenge for the learning process. Even for policy evaluation, following the temporal difference update easily diverges in the linear function approximation \citep{bert_countertd,boyan_safely,gordon_stable,tsi_97}.

\citet{doina_trace_2000} et. al. were the first to use importance sampling for off-policy learning. They used an online updated product of importance sampling ratios to correct the distribution of the behavior policy to that of the target policy, and developed an algorithm that gives a consistent estimation for off-policy evaluation in the lookup table setting. \footnote{There is another class of off-policy learning methods, which does not require importance sampling but to stabilize the underlying ordinary differential equation, e.g., see \citep{gtd,tdc}.} However, importance sampling suffers from high variances, especially when the behavior and target policies are very different. For policy gradient methods, it is hard to see there is a difference. In particular, the importance sampling ratios, widely adopted in recent popular policy gradient algorithms, can be problematic too.

In this context, the Conservative Policy Iteration (CPI) objective~\citep{kakade2002approximately} is a key element in the recent spectrum of popular algorithms, including Trust Region Policy Optimization (TRPO) \citep{schulman2015trust}, Proximal Policy Optimization (PPO) \citep{schulman2017proximal} and many others. CPI is based on the importance sampling ratio between the new and the old policy, which can often cause high variances and lead to poor gradient estimation and unstable performances for policy gradient methods. TRPO avoids this problem by using a fixed threshold for the policy change. PPO applies a clipping method for the importance sampling ratio to ensure it is not too far from the current objective, and then the final objective is the minimum of the clipped and un-clipped objectives. 

A more general topic than off-policy learning is sample-efficient learning. In deep reinforcement learning, sample efficiency is characterized by the following:

1) Efficient policy representation. Ensuring that the updated policy is close to the old policy is a good practice, although there is some approximation error \citep{tomar2021mirror}. The CPI with clipping used by PPO~\citep{schulman2017proximal}, TPPO~\citep{wang2020truly}, and TR-PPO~\citep{wang2019trust} ensures we consider only new policies that are not too far from the old policy. \citet{sun2022you} showed that ratio-regularizer can have a similar effect, and proposed a policy optimization method based on early stopping.
2) Convex optimization. \citet{tomar2021mirror} simplified the problem of maximizing the trajectory's return using convex optimization solvers which minimize the Bregman divergence. 
3) Second-order methods that take advantage of the Hessian matrix. For example, identification of a trust-region is done by computing an approximation to the second-order gradient, such as TRPO~\citep{schulman2015trust} and ACKTR~\citep{wu2017scalable}. However, these methods usually have a high computation cost. 
4) Off-policy learning. Besides the CPI objective-based methods, \citet{wang2016sample} truncated the importance sampling ratio with bias correction and used a stochastic ``dueling network'' to help achieve stable learning. Furthermore, \citet{haarnoja2018soft} proposed an off-policy formulation that reuses of previously collected data for efficiency. Other popular off-policy learning algorithms include soft actor-critic \citep{haarnoja2018soft} and TD3 \citep{fujimoto2018addressing}. 

In this paper, we aim to improve the policy representation due to clipping methods, which provides an improved control of the variances caused by importance sampling. Our goal is achieved by applying the preconditioning technique to the CPI objective, aided with a regularization loss in the policy change. Note that preconditioning is usually applied to an iterative method such as linear system solvers \citep{saad03:IMS,yao2008preconditioned}. Recently,  there has been research on applying preconditioning in deep learning to accelerate the learning process, e.g., see ~\citep{li2016preconditioned,sappl2019deep}. Our work is a new application of preconditioning to control variances in policy gradient estimation. Moreover, our preconditioning technique has an interesting property: it encourages exploration when the policy change is small and switches to exploitation when the policy change is large. 

\section{Background}
Here we review the basis of Markov Decision Processes (MDPs) and recent popular  algorithms TRPO and PPO. 
The key to TRPO and PPO are their objective functions, both of which are based on an approximation to the value function of a new policy.

{\bfseries Markov Decision Processes}.
An MDP is defined by $(\SS,\AA,\PP,R,\lambda)$, where $\SS$ is the state space, $\AA$ is the action space, and for each $a\in \AA$, $\PP$ is a probability measure assigned to a state $s\in \SS$, which we denote as $\PP(\cdot|s,a)$. Define $R: \SS\times \AA \to \RR$ as the reward function, where $\RR$ is the real space. 
$\lambda \in (0,1)$ is the discount factor. Here we consider stochastic policies; denote a stochastic policy by a probability measure $\pi$ applied to a state $s$: $\pi(\cdot|s) \to [0,1]$. 
At a time step $t$,
the agent observes the current state $s_t$ and takes an action $a_t$. The environment provides the agent with the next state $s_{t+1}$ and a scalar reward $R_{t+1}=R(s_t, a_t)$. The main task of the agent is to find an optimal policy that maximizes the expected sum of discounted future rewards:
\begin{gather*}
  V_\pi(s) = \mathbb{E}_{\pi}\Big[\sum_{t=0}^{\infty } \lambda^t R_t \Big],\notag
  \\
  \mbox{where $a_t \sim \pi(\cdot|s_t)$ and $s_{t+1} \sim \PP(\cdot|s_t,a_t)$ for all $t\ge 0$. }
\end{gather*}

The state-action value function for the policy is defined similarly, except the initial action (at $t=0$) is not necessarily chosen according to the policy:
\begin{gather*}
 Q_\pi(s, a) = \mathbb{E}_{\pi}\Big[\sum_{t=0}^{\infty } \lambda^t R_t \Big],\\ \mbox{where $a_t \sim \pi(\cdot|s_t)$ for $t\ge 1$.}
\end{gather*}
We will use $A_{\pi}$ to denote {\em the advantage function}, which can be used to determine the advantage of an action $a$ at a state $s$ by
$A_{\pi}(s, a)=Q_{\pi}(s, a) - V_{\pi}(s)$.
Note that if $a \sim \pi(\cdot|s)$, then the advantage is zero. So this measure computes the advantage of an action with respect to the action that is suggested by the current policy $\pi$.
In the remainder of our paper, $\hat{A}$ is an approximation to the advantage, which is simply the difference between the state-action function and the value function both estimated by a specific algorithm. 

Let $\rho_0$ be the initial state distribution. Let $\eta(\pi)$ be the expected discounted reward:
\[
\eta(\pi) = \mathbb{E}_{s\sim \rho_0}\Big[V_\pi(s)\Big].
\]
Now suppose we are interested in another policy $\tilde{\pi}$. 
Let $d_{\tilde{\pi}}$ be the stationary distribution of the policy. 
According to \citet{kakade2002approximately}, the expected return of  $\tilde{\pi}$ can be calculated in terms of $\eta(\pi)$ and its advantage over $\pi$ in a straight-forward way:
\[
     \eta(\Tilde{\pi}) = \eta(\pi) + \sum_s \rho_{\Tilde{\pi}}(s) \sum_a \Tilde{\pi}(a|s) A_\pi(s,a).
\]
Here $\rho_{\Tilde{\pi}}(s) = \sum_{t=0}^{\infty}\lambda^t d_{\tilde{\pi}}(s_t)$, $s_0 \sim \rho_0$ and the actions are chosen according to $\Tilde{\pi}$, which is just the sum of discounted visitation probabilities. 
\citet{schulman2015trust} approximated $\eta(\Tilde{\pi})$ by replacing $\rho_{\Tilde{\pi}}(s)$ with $\rho_\pi(s)$ in the right-hand side:
\[
    \hat{\eta}_\pi(\tilde{\pi}) = \eta(\pi) + \sum_s \rho_\pi(s) \sum_a \Tilde{\pi}(a|s) A_\pi(s,a).
\]

Note that, in our algorithm, the new policy will be $\tilde{\pi}$. 
The benefit of this approximation is that the expected return of the new policy $\tilde{\pi}$ can be approximated based on the previous samples and the old policy $\pi$. 
Note further, for any parameter value $\theta_0$, because of how $\hat{\eta}$ is defined, we have
\begin{align}
\hat{\eta}_{\pi_{\theta_0}}(\pi_{\theta_0})&=\eta(\pi_{\theta_0}), \label{eq:value_equal_trpo} \\    
\nabla_\theta \hat{\eta}_{\pi_{\theta_0}}(\pi_\theta)|_{\theta=\theta_0}&=\nabla_\theta \eta(\pi_\theta)|_{\theta=\theta_0}.\label{eq:value_equal_trpo_2}
\end{align}
This means that a small gradient ascent update of $\theta_0$ to improve $ \hat{\eta}_{\pi_{\theta_0}}(\pi_{\theta_0})$ also improves $\eta(\pi_{\theta_0})$.

% In practice, $L_\pi(\Tilde{\pi})$\footnote{In the following discussion, to be clear and simplify, for $\theta$ parameterized policy,$L_{\pi_{\theta}}$=$L_{\theta}$,and Unless otherwise specified, in practical algorithm, we omit the subscript$\theta$, just use $L$ represents $L_{\theta}$.} can be estimated using importance sampling.

\paragraph{The TRPO objective.}
% In the TRPO algorithm, the importance sampling method is applied to estimate the objective gradient. For large-scale reinforcement learning tasks, the CPI\citep{kakade2002approximately} objective suffers from infinite variance due to the estimation of trajectory's expected reward by importance sampling. The paper says that there are two general schemes for performing this estimation: single-path and vine. The vine method involves constructing a rollout set and then performing multiple actions from each state in the rollout set. We can use self-normalized importance sampling to reduce the variance of the objective gradient estimation. 
TRPO, PPO and our algorithm P3O all aim to improve a policy incrementally by maximizing the advantage of the new policy over the old one, by considering the influence from importance sampling.  Sample-based TRPO maximizes the following Conservative Policy Iteration (CPI) objective
\begin{equation}
        L^{cpi}(\theta) = \hat{\mathbb{E}}_{t}\left[ r_t(\theta) \hat{A}_{\pi_{old}}(s_t, a_t)\right],\notag
\end{equation}
where $r_t(\theta) = {\pi_{\theta}(a_t|s_t)}/{\pi_{\theta_{old}}(a_t|s_t)}$, by ensuring the difference between the new policy and the old policy is smaller than a threshold. Here the operator $\hat{\mathbb{E}}_{t}$ refers to an empirical average over a finite number of samples.
Note that the importance sampling ratio $r_t(\theta)$ can be very large; 
to avoid this problem, TRPO uses a hard threshold for the policy change instead of a regularization because ``it is difficult to choose a single regularization factor that would work for different problems'', according to \citet{schulman2017proximal}. TRPO then uses the trust region method which is a second-order method that maximizes the objective function with a quadratic approximation. Under that method, the advantage $\hat{A}_{\pi_{old}}$ is replaced by the $Q_{\pi_{old}}$ in TRPO. 

\paragraph{The PPO objective.}
This policy objective function is used to define the PPO algorithm \citep{schulman2017proximal}, which was motivated to augment TRPO with a first-order method extension.
The PPO algorithm first samples a number of trajectories using policy $\pi_{old}$, uniformly  extending each trajectory with $T$ time steps. For each trajectory, the advantage is computed according to 
\[
\hat{A}_t = \lambda^{T-t} V(s_T) + \sum_{k=t}^{T-1} \lambda^{k-t} r_k -V(s_t).
\]
Given the advantages and the importance sampling ratios (to re-weigh the advantages), 
PPO maximizes the following objective:
\begin{equation*}
 L^{ppo}(\theta) = \hat{\mathbb{E}}_t \min\left\{ r_t(\theta) \hat{A}_t, clip(r_t(\theta), 1-\epsilon, 1+\epsilon)\hat{A}_t \right\},
\end{equation*}
The role of the $clip$ operator is to make the policy update more on-policy, and the $min$ operator's role is to optimize the policy out of the clip range. The role of the two composed operators is to prevent the potential instability caused by importance sampling,  and maintain performance across different tasks. Note that this is not the first time that importance sampling causes trouble for reinforcement learning. For the discrete-action problems, both off-policy evaluation and off-policy control are known to suffer from high variances due to the product of a series of such ratios, each of which can be bigger than expected, especially when the behavior policy and the target policy are dissimilar, e.g., see \citep{offpolicy_doina,gtd}.

We noticed some works in literature called PPO an on-policy algorithm. This might be a historical mistake. 
{\em\bfseries The nature of reinforcement learning is truly off-policy}. 
For example, the familiar class of algorithms including Q-learning \citep{watkins1992q}, experience replay \citep{lin1992self}, DQN \citep{mnih2015human}, DDPG \citep{lillicrap2015continuous}, Distributional RL such as C51 \citep{bellemare2017distributional}, QR-DQN \citep{dabney2018distributional} and DLTV \citep{dltv}, Horde\citep{sutton2011horde}, 
Unreal \citep{jaderberg2016reinforcement}, Rainbow \citep{hessel2018rainbow},
LSPI \citep{lagoudakis2003least}, 
LAM-API \citep{yao2012approximate}, Kernel regression MDPs \citep{grunewalder2012modelling} and even MCTS \citep{gelly2011monte}, are all off-policy methods. 

% On-policy methods such as on-policy TD are mostly for studying convergence properties. Sarsa, a small change in the action selection from Q-learning, is an on-policy control method, which may be an exception in this regard. The A3C was motivated to use on-policy methods \citep{mnih2016asynchronous}. However, the exploration and learning by each asynchronous agent is essentially off-policy.  

In off-policy learning, the agent needs to constantly improve its behavior using experience that is imperfect in the sense that it is not learning optimal. Importance sampling ratios arise because one needs to correct the weighting of the objectives from the behavior policy towards the weighting that would be otherwise under a target and improved policy, {\em e.g.}, see \citep{offpolicy_doina,schulman2015trust}. Such re-weighting often leads to high variances in the policy gradient estimations, e.g., see interesting discussions by \citet{truely_policy_gradient}. PPO uses experience from the past (old policies) and also has importance sampling to correct the sample distribution. So clearly PPO is an off-policy algorithm. 

Clipping the importance sampling ratio loses the gradient information of policies in an infinitely large policy space. We define $\tilde{\Pi}_{\epsilon}^+=\{\pi; \frac{\pi(s,a)}{\pi_{old}(s,a)} > 1+\epsilon \}$ and $\tilde{\Pi}_{\epsilon}^-=\{\pi; \frac{\pi(s,a)}{\pi_{old}(s,a)} < 1-\epsilon \}$. Note that PPO’s optimization never crosses into these two spaces for non-negative and negative Advantages, respectively. By clipping, PPO looses the gradient information for any  policy in $\tilde{\Pi}_{\epsilon}^+$ and $\tilde{\Pi}_{\epsilon}^-$. 
We found that there are much better policies within these two policies that PPO fails to discover in our experiments. 
All the PPO algorithms we reviewed extend PPO in some way; however, they unanimously inherit the clipping operation from PPO. 

\section{Method}
In this section, we propose a new objective function by preconditioning for better exploration in the parameter space. And we also use KL divergence to ensure small and smooth policy changes between the updates, balancing exploration and exploitation.

% The approximate objective function needs to meet the following two criteria: 1)the estimated gradient of approximated objective is equal to the original objective in first order at the first mini-batch data; 2) the estimated expected return of sampled trajectories is less than or equal to the ordinary importance sampling estimated return.

% the following questions need to be first addressed:
% \begin{itemize}
%     \item How to achieve a stable estimation of $\nabla_\theta L(\theta)$  considering time and computation?
%     \item  How to ensure a monotonic improvement for policy $\Tilde{\pi}$?
% \end{itemize}

% We divide our method into two parts to solve the infinite variance problem: 1)bound the variance of the gradient estimator using preconditioning; 2)reduce the bias of expected return by minimizing the KL-divergence of the updated policy and target policy.

\begin{figure*}[ht]
\centering
\includegraphics[width=0.90\textwidth]{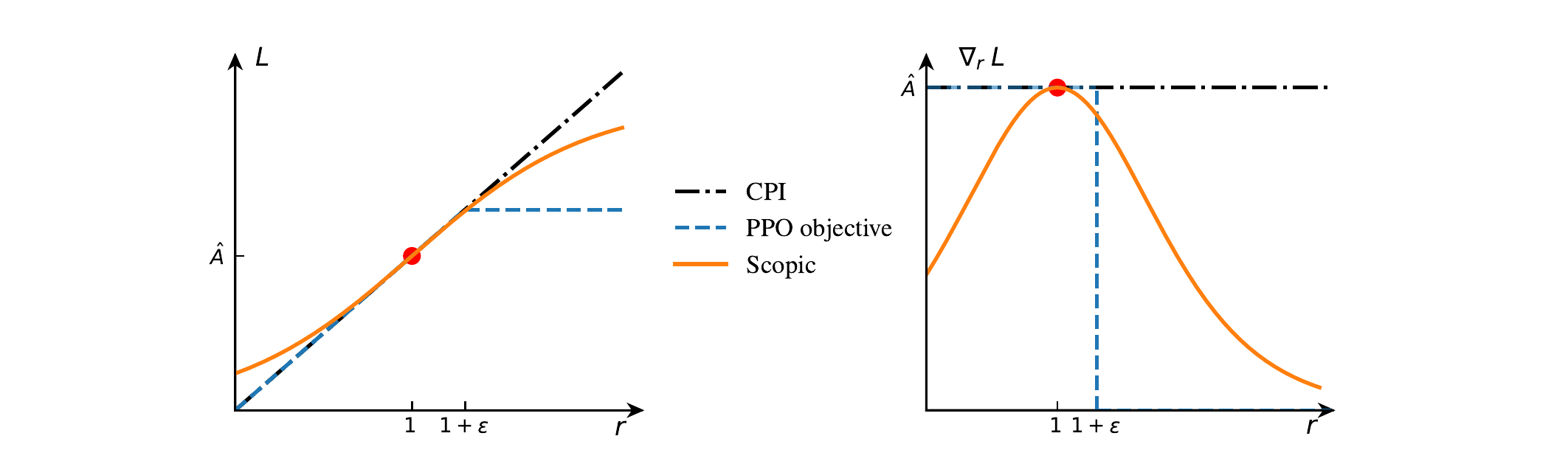} 
\caption{
Left: The objective ($L$) versus the importance sampling ratio ($r$). 
Right: $\nabla_r L$, i.e, the gradient with respect to $r$.  
The $L$ graph comparison includes the CPI objective (black), the PPO objective (blue), and the Scopic objective (orange).  
Both plots are for positive Advantage ($\hat{A}_t>0$) and for a single sample and $\tau=2$.
The red point shows the starting point for the optimization, which is  on-policy learning. 
}
\label{fig:grdients}
\end{figure*} 

\subsection{The Scopic Objective}
Inspired by the value of the  CPI and  PPO objectives, 
% functional knowledge in neural networks~\citep{dugas2009incorporating}, 
we propose the following refined objective:
% \begin{align} \label{eq:psf}
%     J^{sco}({\theta}) = \mathbb{E}_t \left[ \beta_s  \sigma\left(\frac{4}{\beta_s} \left(r_t(\theta) -1 \right)\right) \hat{A}_t\right]
% \end{align}
\begin{align} 
    L^{sc}({\theta}) 
    % = \mathbb{E}_t \left[   \tanh\left(r_t(\theta) -1 \right) \hat{A}_t + \hat{A}_t \right]
    = \hat{\mathbb{E}}_t \Big[ \sigma\left( \tau \left(r_t(\theta) -1 \right)\right) \frac{4}{\tau}\hat{A}_t\Big]
    \label{eq:psf}
\end{align}
where $\sigma$ is the sigmoid function and $\tau$ is the temperature. 
The advantage $\hat{A}_t$ is computed in the same way as in PPO. 
We term this new objective function the {\bfseries Scopic} objective, short for {\bfseries s}igmoidal {\bfseries c}onservative {\bfseries p}olicy {\bfseries i}teration objective without {\bfseries c}lipping.
%\footnote{We also can use the tanh function to construct the sigmoid function, and the tanh function can be converted to and from the sigmoid function, $tanh(x) = 2 \sigma(2x)-1$.}.
Intuitively, according to the Scopic objective, the agent learns to maximize the scaled advantages of the new policy over the old one whilst maintaining stability by feeding the importance sampling ratio to the sigmoid function. Theoretically, by using the sigmoid, the importance sampling ratio is allowed to range from zero to infinity while the output is still in a small range, $[\sigma(-\tau), 1]$. So the new policy is allowed to be optimized in a policy space that is much larger than the clipped surrogate objective. Because the PPO objective is clipped, PPO would not have any information such as the gradient for new policies whose importance sampling ratios over the current policy are beyond the two policy spaces as defined in the PPO objective. 

In addition to constraining the importance sampling ratio range in a ``soft'' way, there is an interesting property of the Scopic objective that is very beneficial for reinforcement learning.
{\em The input of the sigmoid is zero if there is no change in the policy at a state.} Note that the gradient of the sigmoid achieves the maximum in this case. This means when policy change is zero or little, the sigmoid strives for a big parameter update and hence further {\em exploration} in the policy space.
The effect of a big change in $\theta$ leads to a big change in $\pi_\theta$ as well, and thus the action selection has a big change, meaning that our method effectively adapts the parameter update magnitude to explore the action space. 
On the other hand, when the new policy changes greatly from the old policy, the gradient of the sigmoid grows small which gives the parameter little update. The effect is that the agent will focus on a close neighborhood of the policy and use the knowledge built in the policy,
%, repeat similar actions between time steps and collects more samples for the policy,
which leads to {\em exploitation}. So by using the Scopic objective, the agent learns to balance exploration and exploitation automatically via the gradient magnitude that is adapted by the sigmoid function. 
This method of exploration is novel for reinforcement learning and it has not been explored in previous research to the best of our knowledge. 

Existing methods of exploration are mostly based on the novelty of states and actions, typically have some roots in the count-based methods such as UCT \citep{uct} and UCB-1 \citep{auer2002using}. 
The count-based methods have a wide applications in computer games, {\em e.g.}, the use of UCT in AlphaGo  \citep{silver2016mastering}, and the contextual bandits \citep{bubeck2012regret} in recommendation systems  \citep{Li_2010}; {\em etc}. The count-based methods, by definition, only apply to discrete spaces. However, it is possible to extend to some smoothed versions for the continuous case, such as kernel regression UCT \citep{kr_uct}; such methods depend on a choice of kernels and a regression procedure that is performed on a data set of samples. \citet{parameter_exploration} proposed a method that adds Gaussian noise to parameters for exploration and \citet{dltv} discussed parameter uncertainty versus intrinsic uncertainty, and their method implements the UCB principle without counting, by using the distribution information of the value function for uncertainty estimation. In general, the discussion on ``should we be optimistic or pessimistic in the face of uncertainty'' attracts lots of interests from the literature, e.g., see \citep{ciosek2019better,zhang2019quota,keramati2020being,zhou2020non,kuznetsov2020controlling,zhou2021nondecr}.
Our approach via the sigmoidally preconditioned objective is different from the above methods, and it balances exploration and exploitation given an online sample and it does not involve other samples, which is very computationally efficient.   
 
\paragraph{Preconditioning}
At first sight, the Scopic objective term $4/\tau$ may appear odd. Here we explain this choice and what the preconditioner is.
%The $\tau$ is a hyper-parameter to control the boundary whether the expected return and the gradient of the approximate objective are equal to the original objective.
First note the important case when there is no change in the new policy. This momentary on-policy learning can be recovered by $\tau=2$, when the input of the sigmoid function is zero. This means that the learning reduces to that of on-policy, at least for the first mini-batch update. 
Consider the definition of $\hat{\eta}$ in Eq.\ref{eq:value_equal_trpo} and Eq.\ref{eq:value_equal_trpo_2}, for any parameter value $\theta_0$. For the choice of term of $4/\tau$, we can derive
\[
     L^{sc}(\theta_0)= \hat{\eta}(\theta_0), \quad \quad
    \nabla_\theta L^{sc}(\theta)|_{\theta=\theta_0} =\nabla_\theta \hat{\eta}(\theta)|_{\theta=\theta_0}.
    % \mathbb{E}_t[\hat{A} r_t(\theta)]&= \mathbb{E}_t[\hat{A} f(r_t(\theta))].  \label{eq:f_exp}
\]
This ensures that the gradient descent update of the Scopic objective will improve $\hat{\eta}$ and hence $\eta$ for the case of on-policy learning. 

% When $\pi_\theta = \pi_{\theta_{old}}$, then $r_t(\theta)=1$, we have $ \frac{4}{\tau}  \sigma( \tau (r_t(\theta) -1 )) = \frac{2}{\tau}$, so when $\tau=2$, for any policy parameter $\theta_0$, we have
% \begin{equation}
%     L^{sc}(\theta_0)= \hat{\eta}(\theta_0),\label{eq:value_equal}
% \end{equation}
% and the gradient of $L^{sc}(\theta)$ is 
% \[
% E_{t}[ \sigma(\tau (r_t(\theta) -1 )) (1-\sigma(\tau (r_t(\theta) -1 )))  \nabla_\theta r_t(\theta)  4\hat{A}_t ].
% \]

% when $\tau=2$,  $ \sigma(\tau (r_t(\theta) -1 )) (1-\sigma(\tau (r_t(\theta) -1 ))) =\frac{1}{4}$, so, for any policy parameter $\theta_0$, we have
% \begin{equation}
%     \nabla_\theta L^{sc}(\theta)|_{\theta=\theta_0} =\nabla_\theta \hat{\eta}(\theta)|_{\theta=\theta_0}.\label{eq:first_order_equal}
% \end{equation}
% The sigmoid function has the property that $\nabla_x \sigma(x)=\sigma(x)(1-\sigma(x))$.
\begin{figure*}[ht] 
\centering
\includegraphics[width=0.90\textwidth]{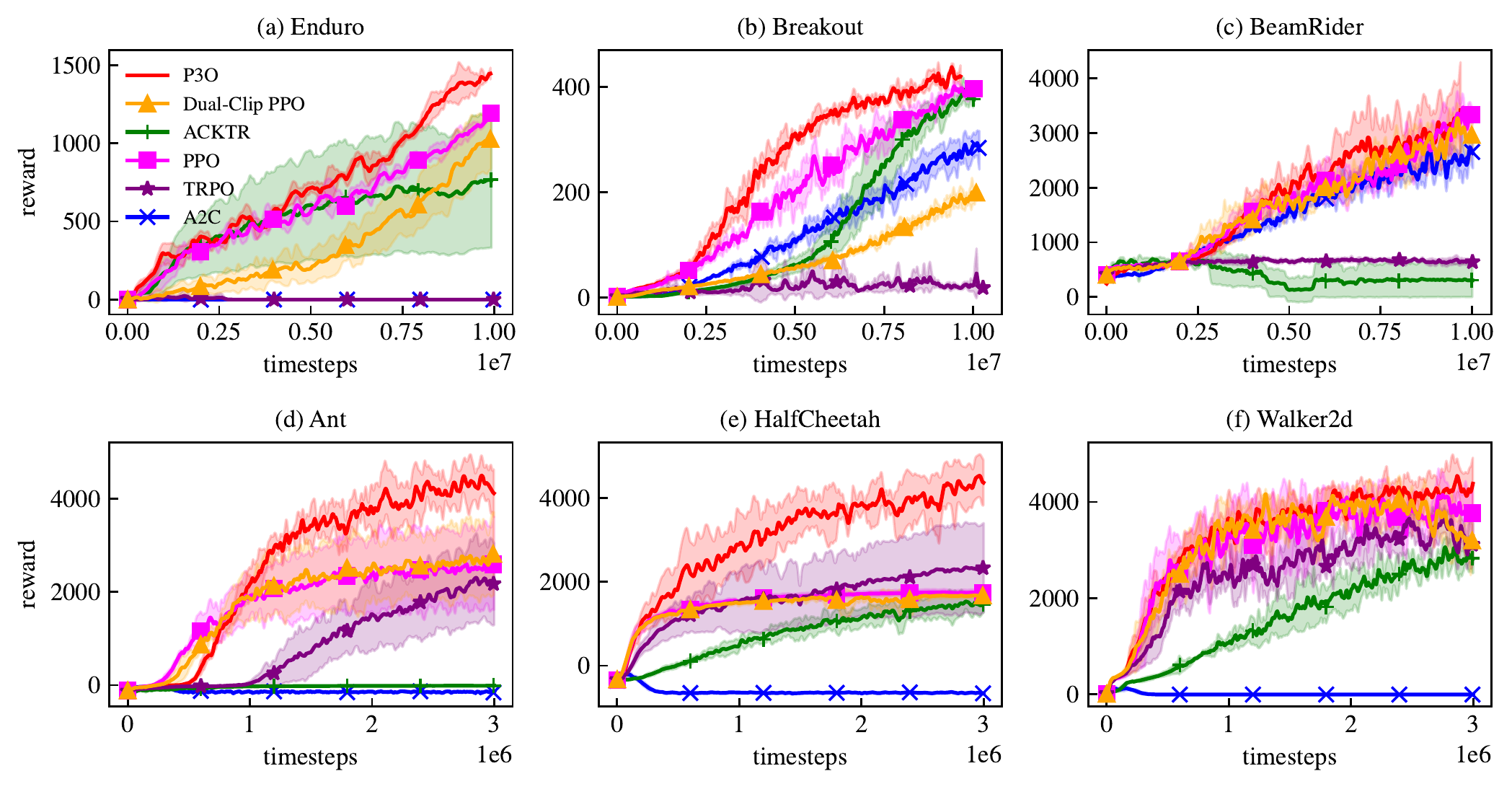} 
\caption{Performance of our P3O versus five baselines for discrete tasks (first row) and continuous tasks (second row).}
\label{fig:performence}
\end{figure*}

The Scopic loss can be viewed as a preconditioning technique. 
Let $p(\theta) = \sigma(\tau (r_t(\theta) -1))$ then the gradients of the Scopic objective and the CPI objective are as follows:
\begin{align}
      \nabla_\theta L^{sc} & = \hat{\mathbb{E}}_{t}[ 4 p(\theta) (1-p(\theta)) \nabla_\theta r_t(\theta)  \hat{A}_t ], \notag
    \\
    \nabla_\theta L^{cpi} &= \hat{\mathbb{E}}_{t}[ \nabla_\theta r_t(\theta)  \hat{A}_t ].\notag
\end{align}

So the stochastic gradient ascent update for the Scopic objective is a modification from that of the CPI objective. This is similar to preconditioning in iterative methods \citep{saad03:IMS}, but note here that the preconditioner is stochastic and applies to the stochastic gradient. %changes the magnitude of the update step but it does not change direction.  
Figure \ref{fig:grdients} shows the objective function and the gradient for CPI, the PPO objective and our Scopic objective. 

\paragraph{Surrogate Function}

The Surrogate function was first introduced in the TRPO paper. The core idea is that when we need to maximize an objective, we can maximize a lower bound instead. We know the PPO algorithm optimizes a proxy function smaller than CPI with a simple analysis. However, from the perspective of optimization, we can view the optimization process of PPO as modifying a one-step on-policy policy gradient to a multi-step mini-batch stochastic optimization, and then using gradient clipping to ensure the stability of optimization. This process includes both on-policy (for the first mini-batch) and off-policy processes.
Our method is also a surrogate function. When $\tau>2$, the Scopic objective is a lower bound of the CPI objective. In experiments, we show that using the sigmoid function is better than gradient clipping for the CPI objective.

\subsection{KL Divergence }\label{sec:KL_rel_MILB}
The Scopic objective function can facilitate more exploration, but sometimes the current policy may only need a little exploration. Therefore, for a better balance between exploration and exploitation, we consider minimizing the KL divergence between the new policy and the old policy. This operation ensures that learning is close to on-policy learning and that the importance sampling ratio will be close to one, especially when the learning rate is decayed. In particular, KL divergence can work well if the current policy is good enough.
Our method has two networks: a policy network and a value network. 
The final objective for our policy network is 
\begin{multline}
    L^{p3o}(\theta)=
    \mathbb{\hat{E}}_t \Big[ \sigma\left( \tau\left( r_t(\theta) - 1\right)\right) \frac{4\hat{A}_t}{\tau} \\-\beta KL\left(\pi_{\theta_{old}}(\cdot|s_t),\pi_{\theta}(\cdot|s_t)\right)\Big],\notag
\end{multline}

where $\beta \ge 0 $ is the regularizer. 
The value networks is trained with the TD(0) algorithm \citep{sutton2018reinforcement},
with the TD update being:
\[
    \hat{\nabla}_w L^{vf}(w)=  \left[r_t+\lambda_v V(s_{t+1}) - V(s_t)\right]\nabla_w V_w(s_t).%\label{eq:v_gradient}
\]
because the objective function is $
    L^{vf}(w)= \hat{\mathbb{E}}_t[r_t+\lambda V_w(s_{t+1}) - V_w(s_t)]$. 

Our Preconditiond Proximal Policy Optimization (P3O) algorithm \footnote{We noted another algorithm also called P3O by \citep{ppo_onpolicy4}.} reduces to the gradient ascent maximizing $L^{p3o}$ and the gradient descent minimizing $L^{vf}$. 
% The P3O algorithm is shown in Appendix~\ref{alg_pa_opt}. 

% \subsection{CLIP is not a good Hyper-parameter to control the information learned from the data}
% All we can see that the TRPO algorithm uses the KL-divergence to control the policy distribution. a smaller KL-divergence means a more conservative policy update. The PPO algorithm uses the CLIP method to achieve a stable policy update.

\section{Empirical Evaluation}
We tested the performance of our P3O algorithm versus baselines in both continuous- and discrete tasks in OpenAI Gym \citep{brockman2016openai} and the Arcade Learning Environment \citep{bellemare2013arcade}. The tasks include Ant-v2, HalfCheetah-v2, and Walker2d-v2 for continuous tasks, for which the  policy is parameterized using  a Gaussian distribution. Discrete tasks include Enduro-v4, Breakout-v4, and BeamRider-v4. The observations on the discrete environments is shown in a four stacking frames RGB image of the screen, and the  policy is parameterized using Softmax distribution.
In addition to TRPO and PPO, we also include A2C \citep{mnih2016asynchronous}, ACKTR \citep{wu2017scalable}, and DualClip-PPO \citep{ye2020mastering} as baselines. 
We evaluate the episodic accumulated reward during the training process of each algorithm. We run each algorithm in the six environments with four random seeds and set the training time steps to be ten million for the discrete tasks and three million for continuous tasks. Both PPO and our P3O do not use augmented data over iterations. Instead, both algorithms use the data from the latest policy.  
% Code is available at https://github.com/raincchio/P3O.
%The details such as the hyper-parameters for all implemented algorithms are contained in Appendix~\ref{ap:parameter}. 
\subsection{Performance Comparison}
% and we provide more detailed results in the Appedix~\ref{sec:more_results}. 
In Figure \ref{fig:performence}, the learning curves of TRPO and A2C are very flat: showing ineffectiveness for these discrete environments.
The TRPO's performance was similar to the empirical results in \citep{wu2017scalable}.
The poor performance may be because it is hard to set a proper parameter for the KL-divergence constraint since it varies in different environments. 
ACKTR is a second-order, natural gradient algorithm with a much higher computational cost per time step. However, it still did not outperform PPO which is a first-order method. 
Dualclip-PPO inherits the clipping operation from PPO objective and adds another max operator with an additional parameter \citep{ye2020mastering}.
The algorithm was applied to the game of Honor of Kings and achieved competitive plays against human professionals. However, there was no baseline comparison. In our experiments, the algorithm performed close to or worse than than PPO. 
That PPO is better than all the other four baselines shows that it is indeed important to control the high variance issue of importance sampling.
Our P3O outperformed all the baselines including the best performing PPO for the tasks.
%This shows the Scopic objective lead to a better performance without clipping. 
In the next subsection, we consider reasons why this happened.

\subsection{The DEON Off-policy Measure and Policy Space Comparison}\label{sec:deon}
\begin{figure}[t]
\centering
\includegraphics[width=0.93\linewidth]{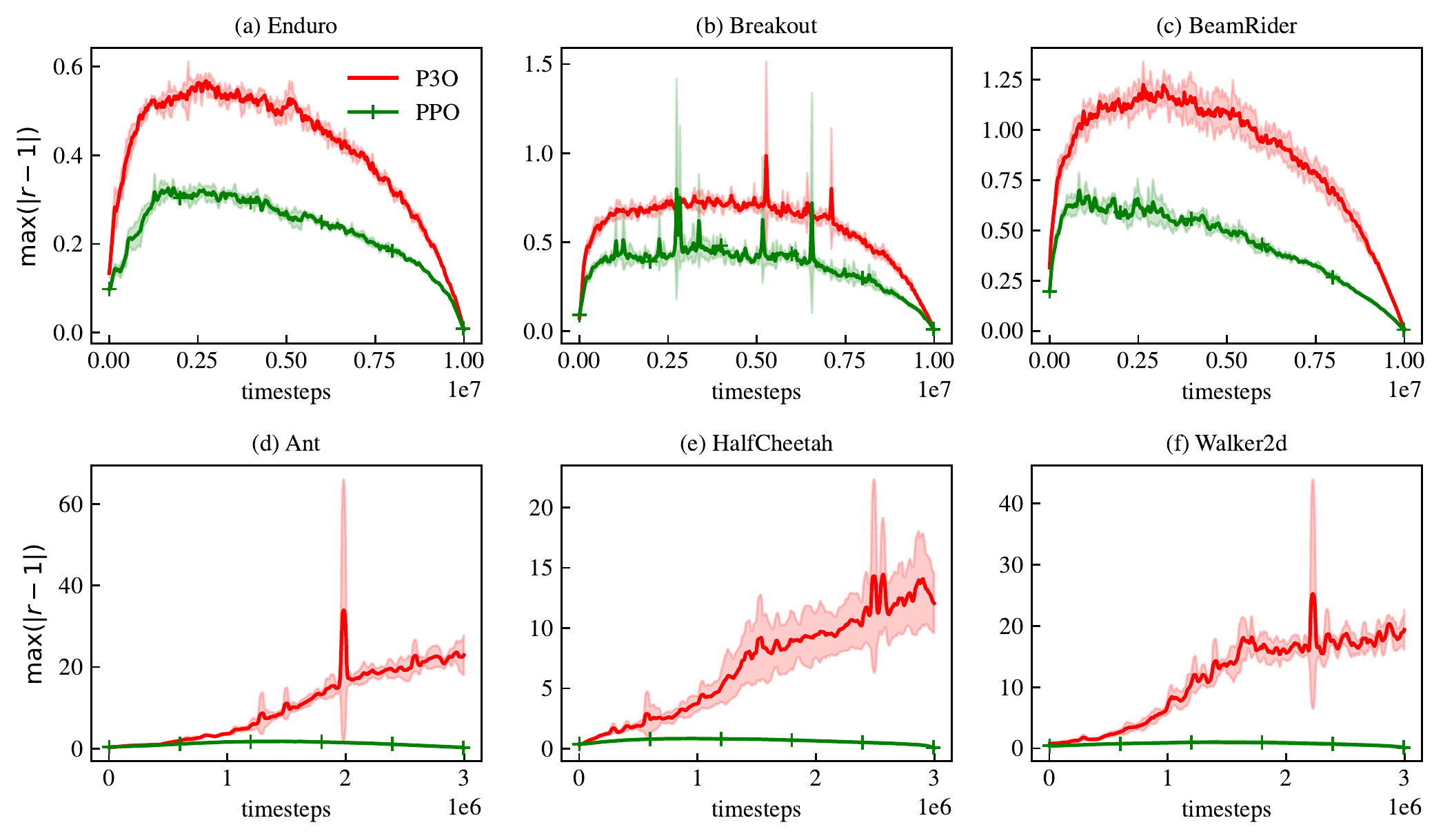} 
\caption{The DEON off-policy measure of our P3O vs. PPO during training. For PPO, the metric is computed without clipping the importance sampling ratio.
Both algorithms use their {\em raw} importance sampling ratios computed during their individual training. 
P3O's deviations are much higher than PPO, which confirms that P3O explores in a policy space that is much bigger than PPO; this is because the clipping in the PPO's objective makes it fail to discover better policies beyond the clip range. 
}
\label{fig:max_ratio}
\end{figure}

In order to understand the performance comparison between PPO and our P3O, we compared the maximum {\em {\bfseries de}viation from {\bfseries on}-policy learning} (DEON) measure,
defined by $y=\max(|r-1|)$, where the maximum was taken over the importance sampling ratio $r$ minus one, absolute, in the collected trajectories of samples. The results of this measure are computed during the training process of PPO and P3O, and compared in Figure \ref{fig:max_ratio}. It shows that the deviation of our P3O from on-policy learning is much bigger than PPO during training: {\em P3O is more off-policy than PPO}. This means P3O explores in a much bigger policy space than PPO. The clipping and minimum operations in the PPO objective prevent the algorithm from exploring the policy spaces $\tilde{\Pi}_{\epsilon}^+$ and $\tilde{\Pi}_{\epsilon}^-$ in the case of non-negative and negative Advantages. 
Together with the performance comparison in Figure \ref{fig:performence}, this shows that the new Scopic objective via sigmoidal preconditioning that is used by our P3O is a very effective way of conducting exploration in the parameter space which results in efficient exploration of the action space.  

Note that, for discrete tasks, all the importance sampling ratios finally converged close to one due to the use of the decaying learning rate. This is interesting because it shows the learning rate decay can give us an on-policy learning algorithm in the long run. For continuous tasks where the fixed learning rate is used, the DEON measure of P3O is still increasing in the end, while for PPO it drops close to zero. 
In particular, the average DEON measure is up to as big as $60.0$ for P3O which performs better than PPO. This shows clipping the importance sampling ratio in the range $[1-\epsilon, 1+\epsilon]$ by the PPO objective is far from sufficient to cover good policies. In our experiments, we set $\epsilon=0.2$, as used in the PPO paper \citep{schulman2017proximal}. 
The DEON metric being still large in the end of learning means P3O is still exploring. The larger policy search space and the consistent exploration leads to a bigger improvement in performance for continuous tasks than for discrete tasks (See Figure \ref{fig:performence} and note the continuous tasks have a much coarser scale in the y-axis). 

Previously, there have been a few measures of policy dissimilarity, especially based on the $L_1$ distance. For example, 
\citet{off-policyness-remi-2} and \citet{off-policyness-remi} used $\norm{\pi(\cdot |s) - \mu(\cdot |s)}_1$ to measure the dissimilarity of the two policies at a state $s$. 
We believe they are also the first to propose the notion of ``off-policyness.'' 
This idea can be also extended to the action dissimilarity at a state \citep{off-policy-Qualitative}, $|\pi(a |s) - \mu(a |s)|$. However, the $L_1$ and  absolute-value based measures are not sufficiently sensitive. Consider two cases, (1) $\pi(a|s)=0.10001, \mu(a|s)=0.00001$; (2) $\pi(a|s)=0.2, \mu(a|s)=0.1$. The dissimilarity according to the absolute-value measure for the two cases is both $0.1$. However, apparently the policies deviate more in the first case. The behaviour policy $\mu$ is a rare event for $a$ at $s$. This suggests that, in the last iteration, we have insufficient samples for the state and action. However, in the second case, there is still a significant percentage of samples. For measuring off-policyness, it appears we need more sensitive metrics. This can be captured well by our DEON measure, which is based on the ratio of the two policies at a state. In particular, the DEON measure is $10000$ for the first case, and $1.0$ for the second case.

\subsection{CPI Objective Comparison}
\begin{figure}[t]
\centering
\includegraphics[width=0.95\linewidth]{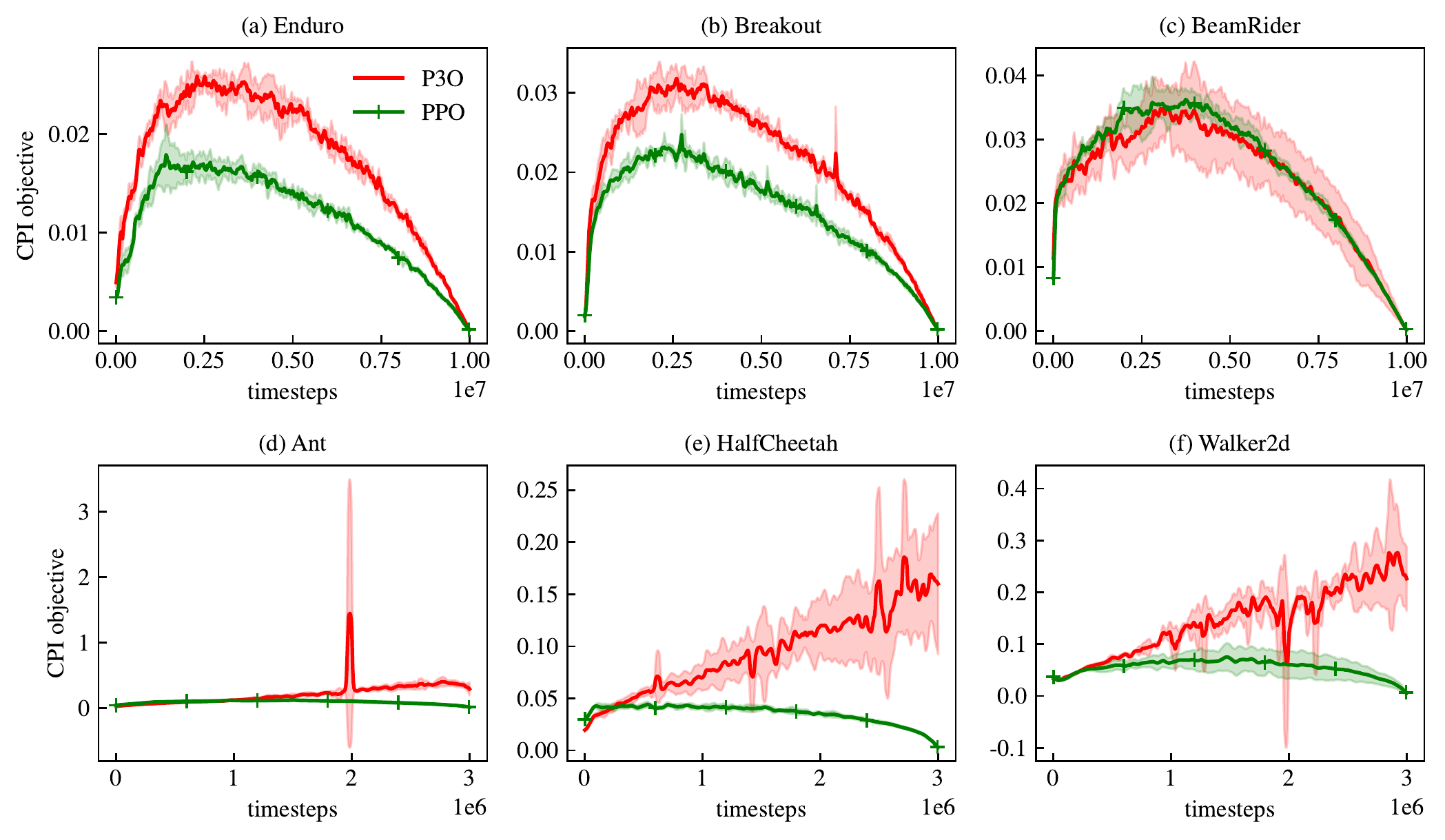} 
\caption{Plotting the CPI objective during training of {P3O} and {PPO}. 
The CPI objective (see the Background Section) has no clipping involved for PPO as well. 
P3O maximizes the CPI objective much better than PPO except for BeamRider.}
\label{fig:cpi_objective}
\end{figure}
This experiment was motivated by noting that both the PPO objective and the Scopic objective originate from the CPI objective. As shown by the TRPO and PPO algorithms, directly maximizing the CPI objective is problematic because of high variances of importance sampling. 
PPO and our P3O can be viewed as special methods for maximizing the CPI objective. So one important question is how much is the CPI objective maximized in either of the two algorithms? We thus calculated the CPI objective (without any clipping or sigmoid preconditioning) in the training process of the algorithms. The result is shown in Figure \ref{fig:cpi_objective}.
For both the discrete and continuous tasks (except BeamRider), P3O consistently maximizes the CPI objective better than PPO. This means the Advantage of the new policy is consistently larger than the old policy with sigmoid preconditioning than with clipping. 
For the discrete tasks, the CPI objective of P3O finally converged close to that of PPO. This is also because the decay learning rate was used for discrete tasks, which became really small in the end. This leads to the importance sampling being one and thus the Scopic objective reducing to the CPI objective. The CPI of continuous tasks is even much bigger. Because of the use of the fixed learning rate, P3O still actively explores even in the end of learning and keeps discovering new policies whose Advantage is much bigger than the old one.

\input{hyper.tex}

\section{Conclusion}
% The clipping operation for importance sampling ratio successfully reduces the variance in policy gradient estimates, creating the popular algorithm PPO. Many improvements of PPO inherit the clipping operation and are optimized in a clipped policy space. 
We proposed a new surrogate objective that applies the sigmoid function to the importance sampling ratio. Our surrogate objective enables us to find much better policies outside of the clipped policy space, and can be viewed as a ``soft clipping'' technique, with a nice exploration property. This extends our understanding of the PPO algorithm and many later developments based on it, and suggests we should look into optimizing the CPI objective beyond the clipped policy space. We found that PPO is insufficient in {\em off-policyness}, and our P3O deviates more from on-policy learning than PPO, according to a measure of off-policyness during training which is called DEON. We can use this metric to measure the policy disparity introduced by the importance sampling method. We compared our P3O algorithm with five recent deep reinforcement learning baselines in both discrete and continuous environments. Results show that our method achieves better performance than the baselines. 

\section{Acknowledgments}
This work is partially supported in part by the National Natural Science Foundation of China under grants Nos.61902145, 61976102, and U19A2065; the International Cooperation Project under grant No. 20220402009GH; and the National Key R\&D Program of China under grants Nos. 2021ZD0112501 and 2021ZD0112502. Dongcui and Bei are partly funded by the Natural Sciences and Engineering Research Council of Canada (NSERC) Discovery grant RGPIN-2022-03034. Bei and Randy are supported by the Alberta 
Machine Intelligence Institute.
%\clearpage
\begin{small}
\bibliography{ref}
\end{small}
\appendix

\section{Algorithm}
The pseudo-code of our P3O algorithm is shown in Algorithm \ref{alg_pa_opt}. 

\begin{algorithm}
\caption{Preconditioned Proximal Policy Optimization}
\label{alg_pa_opt}
\begin{algorithmic}
\STATE {\bfseries Input:}  a simulation environment
\STATE {\bfseries Output:} policy $\pi_{\theta}$ for the environment
\STATE Initialize policy parameters $\theta_{\pi}, \theta_{\pi_{old}}$ and value network parameters $\theta_w$, learning rate for each parameter$\lambda_{\pi}, \lambda_v$ , update number $T$, number of samples $N$, data buffer $\mathbb{D}$
\STATE Simulate as many as possible agents (depending on hardware) with policy $\pi_{\theta}$ to interact with the environment

\FOR{$iteration=1$ {\bfseries to} $m$ }
\REPEAT
\STATE $a_t \sim \pi_\theta(a_t|s_t)$
\STATE $s_{t+1} \sim \PP(s_{t+1}|s_t,a_t)$
\STATE $\mathbb{D} \leftarrow \mathbb{D}\cup {(s_t,a_t,r_t,s_{t+1})}$
\UNTIL{$t \geq NT$}
\STATE Take all the buffer data to compute advantage $\hat{A}$

\STATE $\pi_{\theta_{old}}= \pi_\theta$
\FOR {$epoch=1$ {\bfseries to} $T$}
\STATE Sample $N$ mini-batch samples from Buffer $\mathbb{D}$
\STATE $\theta_\pi \leftarrow\theta_\pi - \lambda_\pi \hat{\nabla}_\theta L^{sc}$
\STATE $\theta_w \leftarrow\theta_w - \lambda_V \hat{\nabla}_w L^{vf} $
\ENDFOR

Clear data buffer $\mathbb{D}$\;

\ENDFOR
\end{algorithmic}
\end{algorithm}

\end{document}

%% file: hyper.tex
\subsection{Sensitivity to Hyper-parameters}
\begin{figure}
\centering
\includegraphics[width=0.99\linewidth]{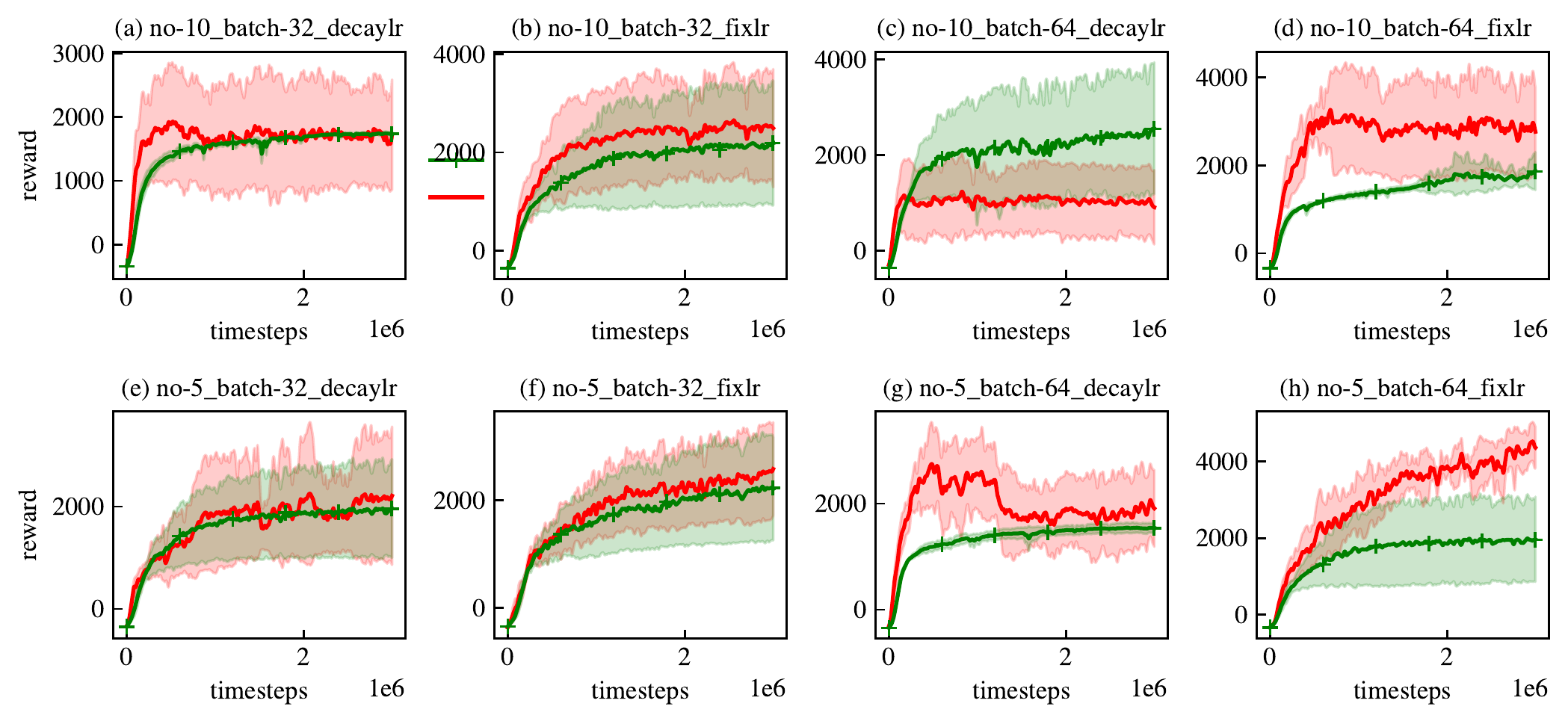} 
\caption{
Hyper-parameter sensitivity studies of PPO (green) and our P3O (red) on HalfCheetah, with two schedulings of each of the three dimensions: the number of epochs on the samples, batch size, and learning rate. In terms of the best case, P3O (group h) performed better than PPO (group c; consistent with the original PPO paper) with higher mean reward and lower variances.  
}
\label{fig:halfcheetah-hyper-parameter}
\end{figure}

The hyper-parameter studies were performed over three dimensions (number of updates, batch size, learning rate). The numbers of epoch updates were either 5 or 10. The batch size were either 32 or 64. 
The learning rate were either constant $10^{-4}$ or the decay scheduling. The decay schedule started with a learning rate of $3\times 10^{-4}$ and decayed linearly, which was used in OpenAI's PPO implementation.  
This leads to eight hyper-parameter combinations whose results are shown in Figure \ref{fig:halfcheetah-hyper-parameter}.
Group (c) performed the best for PPO (consistent with the best result in the PPO paper) and Group (h) was the best for P3O.
This shows PPO prefers the decay schedule while our method prefers the fixed learning rate for continuous tasks. 
For the best hyper-parameter group for both algorithms, P3O's variance is much lower than PPO. 
